\documentclass{article}
\usepackage{spconf,amsmath,graphicx, xcolor,amsfonts}

\title{Trajectory-Prediction with Vision: A Survey}
\name{Apoorv Singh\thanks{This literature review was done while working at Motional.}}
\address{Motional}
\begin{document}
%
\maketitle
\begin{abstract}
To plan a safe and efficient route, an autonomous vehicle should anticipate future trajectories of other agents around it. Trajectory prediction is an extremely challenging task which recently gained a lot of attention in the autonomous vehicle research community. Trajectory-prediction forecasts future state of all the dynamic agents in the scene given their current and past states. A good prediction model can prevent collisions on the road, and hence the ultimate goal for autonomous vehicles: \emph{Collision rate: collisions per Million miles}. The objective of this paper is to provide an overview of the field \emph{trajectory-prediction}. We categorize the relevant algorithms into different classes so that researchers can follow through the trends in the trajectory-prediction research field. Moreover we also touch upon the background knowledge required to formulate a trajectory-prediction problem.
\end{abstract}
\section{Introduction}
\label{sec:intro}
\begin{figure}[htb]
  \centering
  \centerline{\includegraphics[width=8.5cm]{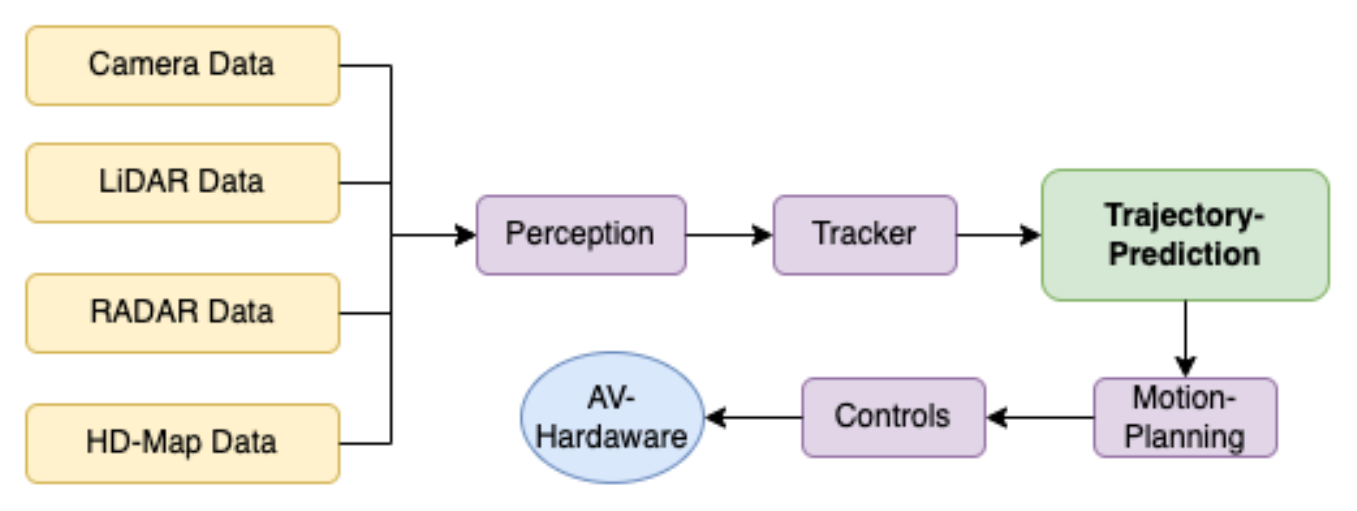}}
\caption{High-level architecture of an Autonomous vehicle stack. In this survey we focus on the green color-coded block. Traditionally, \emph{prediction module} takes input of tracks from the \emph{tracker module}; and outputs tracks with future-trajectories to the \emph{motion-planning module}.}
\label{fig:av-flow}
\end{figure}

Modern approaches to self-driving divide the problem into four steps: detection, object-tracking, trajectory-prediction and path-planning; used in sequence as shown in Fig.\textcolor{blue}{\ref{fig:av-flow}}. In this paper we will go through the trajectory prediction problem which is responsible for forecasting trajectories and intentions of all the dynamic agents on the scene. This module finds its usage in any dynamic robotic platform, self-driving industry being one of the biggest one. To fully solve the trajectory-prediction problem, social intelligence is very important, since we need to bound possibilities, to limit our infinite search space, given our social intelligence. For example: An unattended child at an intersection with pedestrian's red-signal would have more probability to come on the road compared to a full-grown adult in the same situation. \\
\emph{Why is prediction important?} Localization and classification (a.k.a perception) of the objects on the road is one thing, but we also need to understand the dynamics of the agents and their surroundings to predict their future behavior and prevent any crashes. Given the criticality of the problem, self-driving industry has a dedicated module for this on top of the perception and motion-planning module, nowadays. \\
There are several published survey papers on behavior analysis. For example \cite{ijcai_survey} looked at the problem based on the perspective-view approaches, but this coordinate-space is rarely used in the autonomous vehicles industry. All the modeling is done in the Bird's Eye View (BEV) space instead. Historically camera based features were computed in perspective view, but modern techniques \cite{singh2023surround, singh2023vision, mvfusenet} have enabled perception features to be generated in the BEV itself. Hence, it makes more intuitive sense to operate trajectory-prediction module in the BEV space, as is required by its customers viz., motion-planning module shown in Fig. \textcolor{blue}{\ref{fig:av-flow}}. \cite{survey_its} is a little outdated in terms of State-of-the-art (SoTA) methods; with the major focus on HMMs, SVMs, Bayesian based methods et al. A lot of top-performing prediction methods are based on deep-learning-based approaches, which also reflects the current perception models \cite{park20233m3d, singh_tbd, singh_patent_1}. However, to the best of the author's knowledge, this is the first work which covers modern DL-based forecasting approaches targeted for AV-space using all the sensory-data that are available on-board to model accurate prediction models for dynamic agents on the road.

The contributions of this work are summarized as follows:
\begin{itemize}
    \item Paper goes through the basics of the trajectory-prediction problem statement along with all the terminologies around it in  section \textcolor{blue}{\ref{sec:prediction-background}}.
    \item Paper goes through top-trending techniques in trajectory-prediction, primarily focusing on SoTA (State of The Art) methods viz., Transformers and Goal-based models in section \textcolor{blue}{\ref{sec:prediction-methods}}.
    \item Paper provokes researchers with a possible future directions and highlights current research gaps in section \textcolor{blue}{\ref{sec:conclusion}}.
\end{itemize}

\section{Basics and Challenges of Trajectory Prediction}
\label{sec:prediction-background}
In this section we go through the challenges in trajectory-prediction problem statement and some basic background information required to dive deeper into top-trending techniques for the modern trajectory prediction methods.
\subsection{Challenges}
Prediction in the self-driving domain is a complex problem due to the following characteristics:
\begin{itemize}
    \item \textbf{Interdependence:} There are inter-dependencies in the behavior of the agents i.e. future behavior of one agent may affect future behavior of other agents in the vicinity. Hence, we need to take into account the entire surrounding scene of the road including traffic-rules for making agent's trajectory-prediction. This makes prediction modeling as a joint optimization problem for all the agents.
    \item \textbf{Real-time Requirement:} We need to design a bulky deep-learning module that does joint-optimization of the agent's trajectories. However, autonomous vehicles needs to operated in real-time ($\sim$10hz) giving the prediction module a very tight run-time budget.
    \item \textbf{Accumulating Errors:} Prediction module comes after perception and tracking module in self-driving software-stack. This means that there would be some errors already accumulated by the other modules. Hence prediction module's performance is dependent on how well previous models perform.
    \item \textbf{Dynamic Nature:} Both the ego-vehicle and agents are moving in the scene. Future trajectories of the agents depend on the motion of the ego vehicle as well. Hence, ego-vehicle motion compensation needs to be modeled while dealing with temporal data from the sensors. Modeling everything in BEV makes this problem somewhat simpler.
    \item \textbf{Multi-modal Behavior:} Multi-modal behavior of agents, that is given a past history of an agent, there could be multiple possible future trajectories. For example a pedestrian who just stepped on to the cross-walk with a pedesrian red-signal may continue walking or may turn-around. Comprehensive predictor needs to evaluate all the possible trajectories for each event with their likelihood score.
\end{itemize}

\subsection{Prediction task Problem Statement}
\begin{figure}[htb]
  \centering
  \centerline{\includegraphics[width=8.5cm]{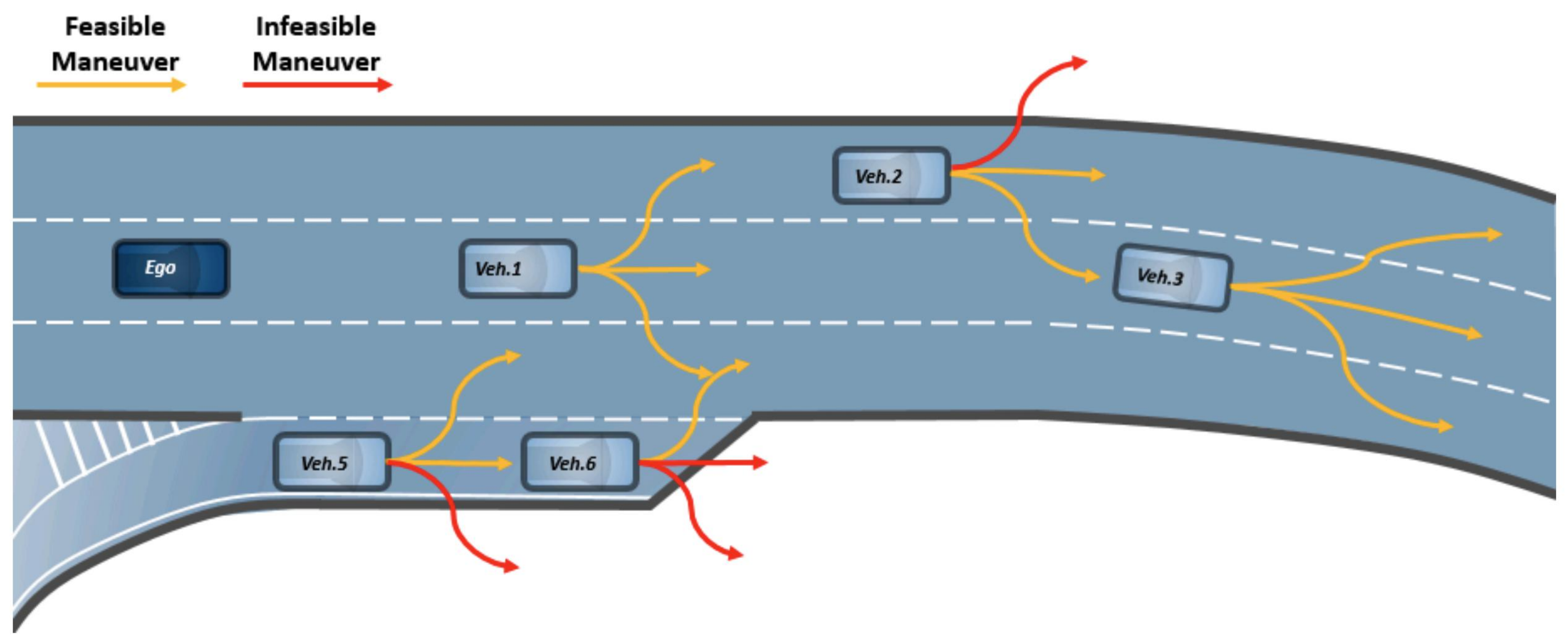}}
\caption{\cite{photo} shows how to model trajectory-predictions for dynamic agents on the road by making them interaction-aware and road-aware. \emph{Veh.2's} trajectory is dependent on \emph{veh.3's} trajectory and vice versa.}
\label{fig:sensor-fusion}
\end{figure}

Prediction task can be divided into two sections as per \cite{ijcai_survey}: 
\begin{itemize}
    \item \textbf{Intention:} This is a classification task where we pre-design a set of intention classes for an agent. For example for a vehicle it could be: \emph{stopped; parked; or moving}. We generally treat it as a supervised learning problem, where we need to annotate the possible classification intents of the agent. 
    \item \textbf{Trajectory:} This division needs to predict a set of possible future locations for an agent in the next $T_{pred}$ future frames, referred as way-points. This constitutes their interaction with other agents as well as with the road as shown in Fig. \textcolor{blue}{\ref{fig:sensor-fusion}}. 
\end{itemize}

Trajectories and intentions need to be interaction-aware. For an instance, it's a fair assumption that an on-coming cars might break a little if you aggressively try to enter a highway on a traffic-packed highway road.
Generally trajectory-prediction can be modeled in image-view (a.k.a perspective-view) or BEV; but now-a-days it is preferred to be done in the BEV space itself. Reason being, in BEV space we can assign a dedicated distance range in the form of grid for our Region of Interest (RoI). However, Image-view can have theoretically infinite RoI because of the vanishing lines in the perspective-view. It is easier to model occlusions in BEV space as motion is more linearly modeled. Ego-motion compensation can be easily done with pose-change (translation and rotation of ego-vehicle) in BEV. Moreover, this space preserves motion and scale of the agents i.e. a vehicle will occupy same number of BEV pixels irrespective of how far it is from the ego-vehicle; which is not the case with image-view.

To predict what will happen in the future, we need to have a good idea of the past. This can be commonly done by using output of the tracker, or it can also be done using historically aggregated BEV features. Goal-based prediction \cite{goal_1} has recently been trending in the literature. This approach argues that in order to predict the future of an agent, we need to have an idea of individual agent's goal.

\subsection{Datasets}
For trajectory-prediction any large-scale perception dataset can be used which includes sequential data viz., nuScenes \cite{nuscenes}; Waymo Open Dataset \cite{waymo}; Lyft \cite{lyft}; ArgoVerse \cite{argoverse}. However, these datasets don't include annotations for intent. LOKI \cite{loki} is a commonly used intention prediction dataset. Trajectory-prediction task can make use of auto-annotations as well if we have sequential unlabeled data; as long we have a good offline perception and tracker model to detect agents and generate temporal link between them.

\subsection{Input-data formats}
Input for prediction can be defined in multiple forms. Simplest way is to send in the sparse tracklets from the tracker. More complex prediction model can feed in BEV representation of the scene; which could be either defined by an occupancy grid or a deep-learning based latent space. 
\subsection{Evaluation Metrics and Losses}
\textbf{Intent Prediction} is a classification task hence Binary cross entropy/ Focal loss \cite{focal} can be used. For evaluation metrics: \emph{precision, recall, F1-score, mean average precision} can be used.
\begin{equation}
Precision=TP/(TP+FP)
\end{equation}
\begin{equation}
Recall=TP/(TP+FN)
\end{equation}
\emph{Key: TP: True Positive; FP: False Positive; FN: False Negative.} \\
\textbf{Trajectory Prediction}: This is a regression problem where we try to regress future way-points of the agent as close as possible to the ground-truth value. For loss computation some version of L1/L2 norm can be used. For evaluation metrics there are multiple ways:
\begin{itemize}
    \item \emph{Final Displacement Error (FDE):} It measures the distance between predicted final location and the true final  location.
\begin{equation}
FDE=|\hat{y}^t_{final} - y^t_{final}|
\end{equation}
\item \emph{Mean Absolute Error (MAE):} It measures the average magnitude of the prediction error based on the root mean square value (RMSE) value. 
\item \emph{Minimum of K Metric:} This is used when $K$ trajectories are predicted per agent by the model. Metric is calculated based on that one of the $K$ trajectories which minimizes the metric error value. 
\end{itemize}
\section{SoTA Prediction Methods}
\label{sec:prediction-methods}
Previous surveys have classified behavior prediction models based on \emph{physics-based}, \emph{maneuver-based} and \emph{interactions-aware models}. \emph{Physics-based} models \cite{ekf-filter} constitute dynamic equations that model hand-designed motions for different categories of agents. This approach fails to model the hidden state of the entire scene and tends to focus only on a particular agent at a time. However, this trend used to be SOTA in pre-deep-learning era. \emph{Maneuver-based models} are the niche models based on the intended motion type of the agents. \emph{Interactions-aware models} are generally an ML based systems \cite{agentformer} that can pair-wise reason every single agent in the scene and produce interaction-aware predictions for all the dynamic agents. There is a high correlation between the trajectories of different agents present nearby in a scene. This approach has shown that it can generalize better by modeling complex attentions module on the agent's trajectories that are heuristically hard to model by a human-designed approach. For rest of the paper we will focus on this category of work. \\
Inspired from \cite{survey_its} we would like to classify trajectory-prediction models based on the input-representations: 
\begin{itemize}
    \item \emph{Tracklets:} A Perception module predicts the current state of all the dynamic agents. This state includes attributes like 3d-center, dimensions, velocity, acceleration etc. Tracker's role is to consume this data and associate it temporally so that each tracklet can contain the state-history for all the agents. Each tracklet now represents the past movement of that agent. This is the simplest form of prediction-model as it contains only sparse tracklets as an input. A good tracker is able to track an agent even if that agent is occluded in the current frame because of its inherent logic. Traditional trackers are non-ML based networks so it can be hard to make the models end-to-end with this approach. A lot of work has been done for interaction-aware modeling with this approach. \cite{track_1} considers tracks of multiple agents in the close-range of the agents with fixed-number of surrounding vehicles for which we are trying to predict the trajectory. They also claim that increasing the receptive field of the tracks helps in improving the behavior prediction performance. \cite{track_2} has looked upon this problem in distance-terms for shortlisting all the agents by specifying a distance-range around the target agent. \cite{track_3} adds attention modeling for different agent classes together with their respective weights.
    \item \emph{BEV Representation:} This branch of networks generates a BEV representation of the current perception output, history of perception outputs and road state with the information from the HD-map all in a single-stacked BEV representation which is consumed by the prediction model. \cite{bev_1} uses camera and RADAR to model BEV representation as an occupancy grid. Dynamic Occupancy Grid Map (DOGMa) \cite{dogma} is created from the data fusion of a variety of sensors and provides a BEV image of the environment. Channels of this grid contains probability of the occupancy and velocity vector for each cell in the grid. As a drawback, this category still suffers through the problem of accumulation of errors; form perception and tracker module.
    \item \emph{Raw-sensor Data:} This is an end-to-end approach where the model takes in raw-sensor data information and directly predicts trajectory-prediction for each agent in the scene. This approach may or may not have auxiliary outputs and their losses to supervise the complex training as in \cite{pnp, fnf}. \cite{v2v} extends the previous approach by adding collaborative perception-prediction for multiple AVs on the road with vehicle-to-vehicle (V2V) communications platform. Drawback of this category is that it is computationally expensive due to dense information for input. Also since it combines three problems together i.e. perception, tracking and prediction; model becomes very hard to develop and even harder to converge.
\end{itemize}

\textbf{Goal-based Prediction} Along with the scene context, the behavioral intention prediction is commonly influenced by different intended goals, and should be inferred with interpretablity. For the goal-conditioned future prediction, the \emph{goal} is modeled as the future state (defined as destination coordinates) \cite{goal_1} or moving types that an agent desires \cite{goal_2}. To break down this problem into two categories: The first one is \emph{epistemic}: to answer the question \emph{Where are the agents
going?} The second is \emph{aleatoric} to answer the question \emph{how is this agent going to reach its goal?}
\section{Further Extensions and Conclusion}
\label{sec:conclusion}
With perception tasks being solved by more than 90\% on their respective metrics and planning tasks being considered to be easily solved as long as accurate predictions are provided; solving trajectory-prediction remains one of the key challenges to unleash autonomous vehicles at the large-scale. Accurate trajectory modeling is the key to prevent possible future collisions; it also needs to run in real-time so that collision avoidance action can be taken within physical limits and reasonable-comfort-level of the users.  In this work, we make a thorough review of existing methods on trajectory prediction for self-driving. We divide different approaches in various classes to easily follow through the trends. Moreover, we went through the trajectory-prediction basics so that readers can easily follow through the paper. We hope we are able to highlight key concerns and trends with this survey paper, which can provoke further research in the field. \\

\bibliographystyle{IEEEbib}
\bibliography{main}

\end{document}